\documentclass[letterpaper, 10 pt, conference]{ieeeconf}  

\IEEEoverridecommandlockouts                              

\usepackage{color}
\usepackage{times}
\usepackage[pdftex]{graphicx}
\usepackage{subfigure}
\usepackage{amsmath,amssymb,amsopn,amstext,amsfonts}
\usepackage{cancel}
\usepackage[space]{cite}
\usepackage{pdfsync}
\usepackage{balance}
\usepackage{color}
\usepackage{mathtools}
\usepackage{algpseudocode}
\usepackage{footmisc}
\usepackage[ruled,vlined,linesnumbered]{algorithm2e}

\algnewcommand\algorithmicforeach{\textbf{for each}}
\algdef{S}[FOR]{ForEach}[1]{\algorithmicforeach\ #1\ \algorithmicdo}
\usepackage{bm}

\usepackage{diagbox}
\usepackage{epstopdf}
\usepackage{pifont}
\usepackage{fixltx2e}
\usepackage{amsmath}
\usepackage{multirow}
\usepackage{url}
\usepackage{verbatim}
\usepackage[linkcolor=black,citecolor=black,urlcolor=black,colorlinks=true]{hyperref}
\usepackage[skip=2pt, font={small}]{caption} 

\title{\LARGE \bf
	A fast, complete, point cloud based loop closure for LiDAR odometry and mapping
}

\author
{Jiarong Lin and Fu Zhang 
	\thanks{J. Lin and F. Zhang are with the Department of Mechanical Engineering, Hong Kong University, Hong Kong SAR., China. {\tt\small $\{$jiarong.lin,  fuzhang$\}$@hku.hk}}
}

\begin{document}
	\maketitle
	\thispagestyle{empty}
	\pagestyle{empty}
	
	\newcommand{\todo}[1]{\textcolor{red}{\emph{\bf#1}}}
	\newcommand{\note}[1]{\textcolor{red}{\emph{\bf#1}}}
	\newcommand{\transpose}{\mbox{${}^{\text{T}}$}}
	\newcommand{\eye}[1][]{\ensuremath{\mathbb{I}_{#1}}}
	\newcommand\footnoteref[1]{\protected@xdef\@thefnmark{\ref{#1}}\@footnotemark}

	\newlength{\bibitemsep}\setlength{\bibitemsep}{.0238\baselineskip}
	\newlength{\bibparskip}\setlength{\bibparskip}{0pt}
	\let\oldthebibliography\thebibliography
	\renewcommand\thebibliography[1]{%
		\oldthebibliography{#1}%
		\setlength{\parskip}{\bibitemsep}%
		\setlength{\itemsep}{\bibparskip}%
	}
	
	\begin{abstract}
		This paper presents a  loop closure method to correct the long-term drift in LiDAR odometry and mapping (LOAM). Our proposed method computes the 2D histogram of keyframes, a local map patch, and uses the normalized cross-correlation of the 2D histograms as the similarity metric between the current keyframe and those in the map. We show that this method is fast, invariant to rotation, and produces reliable and accurate loop detection.  The proposed method is implemented with careful engineering and integrated into the LOAM algorithm, forming a complete and practical system ready to use. To benefit the community by serving a benchmark for loop closure, the entire system is made open source on Github \footnote{\url{https://github.com/hku-mars/loam_livox}\label{loam_livox}}.
	\end{abstract}
	
	\section{Introduction}
	
	With the capacity of estimating the 6 degrees of freedom (DOF) state, and meanwhile building the high precision maps of surrounding environments, SLAM methods using LiDAR sensors have been regarded as an accurate and reliable way for robotic perception. In the past years, LiDAR odometry and mapping (LOAM) have been successfully applied in the field of robotics, like self-driving car\cite{levinson2011towards}, autonomous drone \cite{bry2012state, gao2019flying}, field robots surveying and mapping\cite{nuchter20076d,schwarz2010lidar}, etc. In this paper, we focus on the problem of developing a fast and complete loop closure system for laser-based SLAM systems. 
	
	Loop closure is an essential part of SLAM system, to estimate the long term accumulating drift caused by local feature matching. In a common paradigm of loop closure, the successful detection of loops plays an essential role. Loop detection is the ability of recognizing the previously visited places, by giving a measurement of similarity between any two places. For visual-slam methods, the loop closure considerably benefits from various largely available computer vision algorithms. For example, by utilizing the bag-of-words model \cite{galvez2012bags,filliat2007visual} and clustering the feature descriptors as words, the similarity between observations can be computed in the word space. This kind of method has been used in most of the state of the art visual SLAM system (e.g. \cite{mur2017orb, qin2018vins}) and have achieved great success in the past years.
	
	Unlike the visual-SLAM, the relevant research of laser-based loop closure is rare, and it is surprisingly hard for us to find any available open soured solution which addresses this problem. We conclude these phenomenons as two main reasons: Firstly, compared to the camera sensors, the cost of LiDAR sensors are extremely expensive, preventing them in wider use. In most of the robotics navigation perception, LiDARs is not always the first choice. Secondly, the problem of place recognition on point cloud is very challenging. Unlike 2D images containing rich information such as textures and colors, the available informations in point cloud are only the geometry shapes in 3D space.
	
\begin{figure}[t]
		\centering
		\includegraphics[width=1.0\linewidth]{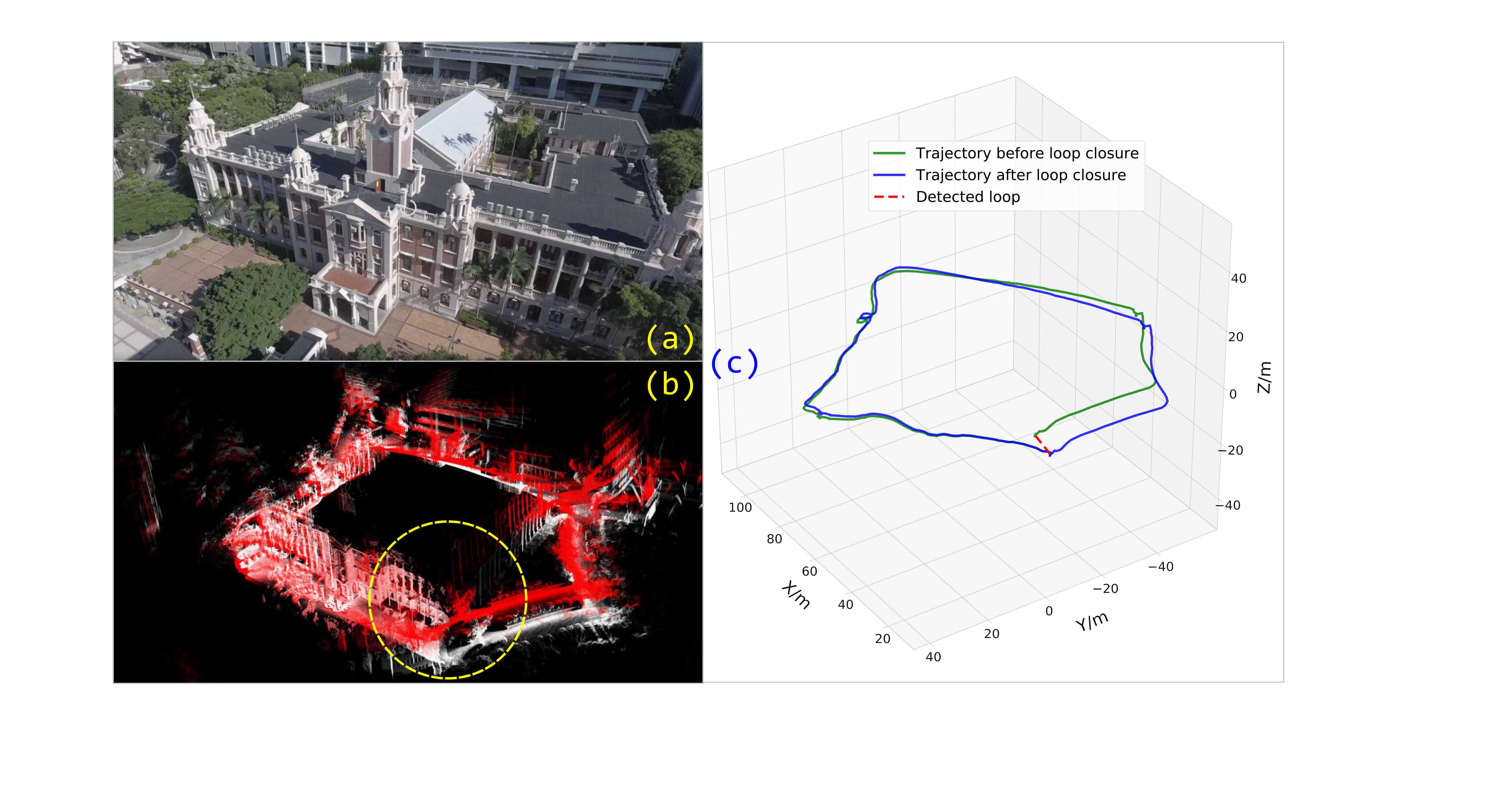}
		\caption{ An example of loop closure around the Main Building of the Hong Kong University (HKU).  (a), the RGB image of the map area; (b), the red and withe points are off the map before and after loop closure, respectively; (c), the red dashed line indicates the detected loop, the green, and blue solid lines are the trajectories before and after loop closure, respectively.}
		\label{fig_merge_hku_zym}
		\includegraphics[width=1.0\linewidth]{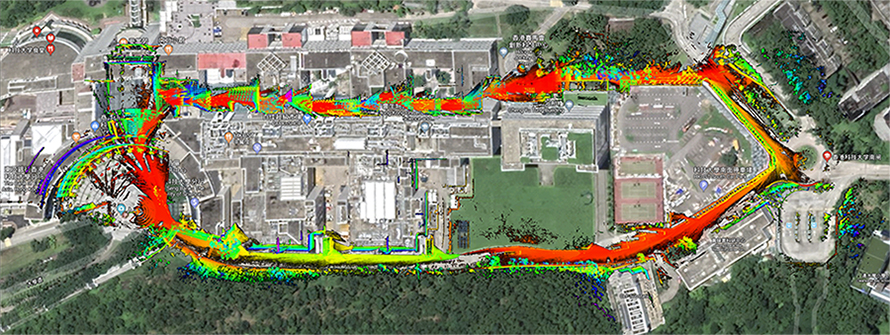}
		\caption{The large scale loop closure of the Hong Kong University of Science and Technology (HKUST) campus. We align the point cloud map after loop closure with the satellite image. Our video is available at \url{https://youtu.be/fOSTJ_yLhFM} .}
		\label{fig_merge_hkust}
		\vspace{-1.5cm}
	\end{figure}

	In this paper, we develop a fast, complete loop closure system for LiDAR odometry and mapping (LOAM), consisting of fast loop detection, maps alignment, and pose graph optimization. We integrate the proposed loop closure method into a LOAM algorithm with Livox MID40\footnote{\label{mid40}\url{ https://www.livoxtech.com/mid-40-and-mid-100}} sensor, a high performance low cost LiDAR sensor easily available. Some of the results we obtain are shown in Fig. \ref{fig_merge_hku_zym} and Fig. \ref{fig_merge_hkust}. To contribute to the development of laser-based slam methods, we will open source all the datasets and codes on Github\footref{loam_livox}.
	

	\begin{figure*}[htp]
		\centering
		{\includegraphics[width=2.0\columnwidth]{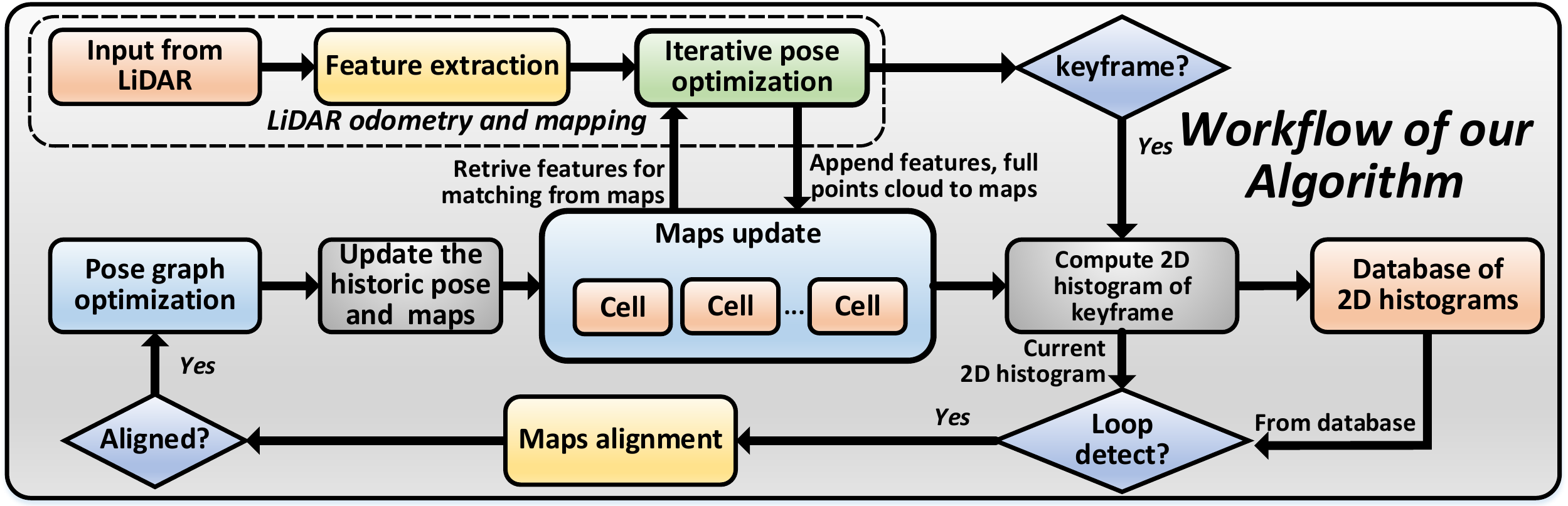}}
		\caption{The overview of our system.}	
		\label{fig_workflow}
		\vspace{-0.2cm}
	\end{figure*}
	
	\section{Related work}
	
	Loop closure is widely found in visual-SLAM to correct its  long-term drift. The commonly used pipeline mainly consists of three steps: First, local feature of a 2D images is extracted by using handcrafted descriptors such as Scale-Invariant Feature Transforms (SIFT) \cite{lowe1999object},  Binary Robust Independent Elementary Features (BRIEF) \cite{calonder2010brief},  Oriented Fast and Rotated BRIEF(ORB) \cite{rublee2011orb}, etc. Then,  a bag-of-world model \cite{galvez2012bags,filliat2007visual} is used to cluster these features and build a dictionary to achieve loop detection. Finally, with the detected loop, a pose graph optimization is formulated and solved update the historic poses in the map.
	
	Unlike the visual-SLAM, loop detection for point cloud data is still an open, challenging problem in laser-based SLAM. Bosse \textit{et al} \cite{bosse2013place} achieve place recognition by directly extracting keypoints from the 3D point cloud and describe them with a handcrafted \textit{3D Gestalt} descriptors. Then the keypoints vote for their nearest points and the scores are used to determine if a loop is detected. The similar method is also used in \cite{gawel2016structure}. Magnusson \textit{et al} \cite{magnusson2009appearance} describe the appearance of 3D point cloud by utilizing the normal distribution transform (NDT),  they compute the similarity of two scans from the histogram of the NDT descriptors.
	
	Besides the hand-crafted descriptors, learning-based method has also been used in loop detection(or place recognitions) in recent years. For example, the SegMatch proposed by Dube \textit{ et al} \cite{dube2017segmatch} achieves place recognition by matching semantic features like buildings, tree, vehicles, etc. Angelina \textit{et al} \cite{angelina2018pointnetvlad} realize place recognition by extracting the global descriptor from an end-to-end network, which is trained by combining the PointNet \cite{qi2017pointnet} and NetVLAD \cite{arandjelovic2016netvlad}. The learning-based method is usually computationally expensive, and the performances greatly rely on the dataset in the training process.
	
	
	Although with these reviewed work, to our best knowledge, there is no open-sourced codes or dataset that benchmark the problem of loop closure for LOAM, which leaves some difficulties for readers on reproducing their works.  To this purpose, we propose a complete loop closure system. The loop detection of our work is mainly inspired by the method of \cite{magnusson2009appearance} and some of the adjustments are made in our scenarios.
	Due to the small FoV and special scanning pattern of our considered LiDAR sensor, we perform loop detection for an accumulated time of scans (i.e., the {\it keyframe}). To summarize, our contributions are threefold: (1) we develop a fast loop detection method to quickly measure the similarity of two keyframes; (2) we integrate our loop closure system, consisting of the loop detection, map alignment, and pose optimization into an LiDAR odometry and mapping algorithm (LOAM) \cite{lin2019loam}, leading to a complete and practical system ready to use; (3) we provide an available solution and paradigm for point cloud based loop closure by opening source our systems and datasets on Github.
	
	\section{System Overview}
	The workflow of our system is shown in Fig.~\ref{fig_workflow}, each new frame (i.e., scan) input from LiDAR is registered to the global map by LOAM algorithm \cite{lin2019loam}. If a specified number of frames have been received (e.g., 100 frames),  a keyframe created, which forms a small local map patch. The raw points, which were registered to the cells of the global map (Section IV) by LOAM, corresponding to the new keyframe are retrieved to compute its 2D histogram (see Section V). The computed 2D histogram is compared to the database, which contains 2D histograms of the global map consisting of all the past keyframes, to detect a possible loop (see Section VI). Meanwhile, the new keyframe 2D histogram is added to the database for the use of next keyframe. Once a loop is detected, the keyframe is aligned with global map and a pose graph optimization is performed to correct the drift in the global map.

	\section{Map and cell}
	In this section, we will introduce two key elements of our algorithm, the maps and cell. For conveniently, We use $\boldsymbol{\mathcal{M}}$ and  $\boldsymbol{\mathcal{C}}$ denote the map and cell, respectively. 
	\subsection{Cell}
	A cell is a small cube of a fixed size (i.e., $S_x, S_y$ and $S_z$ in $x$, $y$ and $z$ directions) by partitioning the 3D space.  It is represented by its center location $\boldsymbol{\mathcal{C}}_c$ and created by the first point $\mathbf{P}_i = [\mathbf{P}_{i_x},\mathbf{P}_{i_y},\mathbf{P}_{i_z}]^T $ in it
	\begin{equation}\label{eq_cell_center}
	\boldsymbol{\mathcal{C}}_c = 
	\begin{bmatrix}
	\text{round}(\mathbf{P}_{i_x} / S_x)*S_x + S_x/2\\
	\text{round}(\mathbf{P}_{i_y} / S_y)*S_y + S_y/2\\
	\text{round}(\mathbf{P}_{i_z} / S_z)*S_z + S_z/2
	\end{bmatrix}
	\end{equation}
	
	Let $N$ denote the number of points located in a cell $\boldsymbol{\mathcal{C}}_c$, the mean $\boldsymbol{\mathcal{C}}_\mu$ and covariance $\boldsymbol{\mathcal{C}}_{\Sigma}$ of this cell is:
	\begin{align}
	\boldsymbol{\mathcal{C}}_\mu &= \dfrac{1}{N} \left( \sum_{i = 1}^{N}{\mathbf{P}_i} \right) \label{eq_cell_mean}\\
	\boldsymbol{\mathcal{C}}_{\Sigma} &= \dfrac{1}{N-1} \left( \sum_{i = 1}^{N}{
		\left(\mathbf{P}_i - \boldsymbol{\mathcal{C}}_\mu\right)\left(\mathbf{P}_i - \boldsymbol{\mathcal{C}}_\mu\right)^T
	} \right) \label{eq_cell_cov}
	\end{align}
	
	Notice that the cell is a fixed partitioning of the 3D space are is constantly populated with new points. To speed up the computation of mean \ref{eq_cell_mean} and covariance \ref{eq_cell_cov}, we derive its recursive form as follows. Denote $\mathbf{P}_{N+1}$ the new point, $N$ is the number of existing points in a cell with mean $\boldsymbol{\mathcal{C}}_{\mu}'$ and covariance $\boldsymbol{\mathcal{C}}_{\Sigma}'$. The mean $\boldsymbol{\mathcal{C}}_{\mu} $ and covariance $\boldsymbol{\mathcal{C}}_{\Sigma}$  of all the $N+1$ points in the cell are:
	\begin{eqnarray}
	\label{eq_cell_mean_incre}
	\boldsymbol{\mathcal{C}}_{\mu} &=& \dfrac{1}{N+1}( N \boldsymbol{\mathcal{C}}_{\mu}' + \mathbf{P}_{N+1}) 
	\end{eqnarray}
	\begin{equation}
	\begin{aligned} 
	\boldsymbol{\mathcal{C}}_{\Sigma} &= \dfrac{1}{N}\sum_{i=1}^{N+1}{ (\mathbf{P}_i-\boldsymbol{\mathcal{C}}_{\mu}) (\mathbf{P}_i-\boldsymbol{\mathcal{C}}_{\mu})^T} \\
	&= \dfrac{1}{N}\sum_{i=1}^{N+1}{ ( \mathbf{P}_i- \boldsymbol{\mathcal{C}}_{\mu}' + \boldsymbol{\mathcal{C}}_{\mu}' - \boldsymbol{\mathcal{C}}_{\mu} ) ( \mathbf{P}_i- \boldsymbol{\mathcal{C}}_{\mu}' + \boldsymbol{\mathcal{C}}_{\mu}' - \boldsymbol{\mathcal{C}}_{\mu} )^T } \\
	&=\dfrac{1}{N}\bigg[ (N-1)\boldsymbol{\mathcal{C}}_{\Sigma}' + (\mathbf{P}_{N+1}- \boldsymbol{\mathcal{C}}_{\mu}')(\mathbf{P}_{N+1}-\boldsymbol{\mathcal{C}}_{\mu}')^T \bigg.\\ 
	&~~~~ \left. + (N+1)( \boldsymbol{\mathcal{C}}_{\mu}' - \boldsymbol{\mathcal{C}}_{\mu} )(\boldsymbol{\mathcal{C}}_{\mu}' - \boldsymbol{\mathcal{C}}_{\mu})^T  \right.\\
	&~~~~ \bigg. + 2\left( \boldsymbol{\mathcal{C}}_{\mu}' - \boldsymbol{\mathcal{C}}_{\mu} \right)(\mathbf{P}_{N+1} - \boldsymbol{\mathcal{C}}_{\mu}')^T  \bigg]
	\end{aligned}
	\label{eq_cell_cov_incre}
	\end{equation}
	
	Therefore, a cell $\boldsymbol{\mathcal{C}}$ is composed of its static center $\boldsymbol{\mathcal{C}}_{c}$, the dynamically updated mean $\boldsymbol{\mathcal{C}}_{\mu}$ and covariance $\boldsymbol{\mathcal{C}}_{\Sigma}$, and the raw points collection $\{ \mathbf{P}_{i} \}$:  $\boldsymbol{\mathcal{C}} = (\boldsymbol{\mathcal{C}}_{c}, \boldsymbol{\mathcal{C}}_{\mu}, \boldsymbol{\mathcal{C}}_{\Sigma}, \{ \mathbf{P}_{i} \} )$.

	\subsection{Map}
	
	The map $\boldsymbol{\mathcal{M}}$ is the collection of all raw points saved in cells. More specifically,  $\boldsymbol{\mathcal{M}}$ consists of a hash table $\boldsymbol{\mathcal{H}}$ and a global octree $\boldsymbol{\mathcal{O}}$. The hash table $\boldsymbol{\mathcal{H}}$ enables to quickly find the specific cell according to its center $\boldsymbol{\mathcal{C}}_c$. The octree $\boldsymbol{\mathcal{O}}$ enables to find out all cells located in the specific area of given range. These two are of significant importance in speeding up the maps alignments.

	For any new added cell $\boldsymbol{\mathcal{C}}$, we compute its hash index $\mathcal{H}(\boldsymbol{\mathcal{C}}_c)$ using the XOR operation of hash index of its individual components: ($\mathcal{C}_{c_x}, \mathcal{C}_{c_y}, \mathcal{C}_{c_z}$). The computed hash index is then added to the hash table of the map $\boldsymbol{\mathcal{H}}$. Since the cell is a fixed partitioning of the 3D space, its center location $\boldsymbol{\mathcal{C}}_c$ is static, requiring no update for existing entries in the hash table (the hash table is dynamically growing though). 
	
	The new added cell $\boldsymbol{\mathcal{C}}$ is also added to the Octree $\boldsymbol{\mathcal{O}}$ according to its center location, similar to the OctoMap in \cite{hornung2013octomap}. Algorithm~\ref{alg_reg_new_scan} illustrates the procedure of incrementally creating cells and maps from new frames. 
	
	\begin{algorithm}[t]
		\caption{Registration of new frame  }
		\label{alg_reg_new_scan}
		\renewcommand{\thealgocf}{}
		\renewcommand{\theAlgoLine}{}  
		\SetKwInOut{Input}{Input}
		\SetKwInOut{Output}{Output}
		\SetKwInOut{Begin}{Begin}
		\SetKwProg{Fn}{Function}{}\\
		\Input{Points $ \boldsymbol{\mathcal{P}}_k $ from $k$-th frame, Current map  $\boldsymbol{\mathcal{M}}$, the pose $\left( \mathbf{R}_k, \mathbf{T}_k \right)$ estimated from LOAM algorithm}
		\For{each $\mathbf{P}_l \in \boldsymbol{\mathcal{P}}_k$}
		{
			Transform $\mathbf{P}_l$ to global frame by $\mathbf{P}_i = \mathbf{R}_k \mathbf{P}_l + \mathbf{T}_k $. \\
			Compute the cell center $\boldsymbol{\mathcal{C}}_c$ from (\ref{eq_cell_center}).\\
			Compute the hash index $\mathcal{H}(\boldsymbol{\mathcal{C}}_c)$.\\
			\If{ $\mathcal{H}(\boldsymbol{\mathcal{C}}_c) \notin \boldsymbol{\mathcal{H}}$ }
			{
				Create new cell $\boldsymbol{\mathcal{C}}$ with center $\boldsymbol{\mathcal{C}}_c$.\\
				Insert $\mathcal{H}(\boldsymbol{\mathcal{C}}_c) $ to hash table $\boldsymbol{\mathcal{H}}$ of map $\boldsymbol{\mathcal{M}}$. \\
				Insert $\boldsymbol{\mathcal{C}}_c$ to \textit{Octotree} $\boldsymbol{\mathcal{O}}$ of map $\boldsymbol{\mathcal{M}}$.
			}
			Add $\mathbf{P}_i$ to $\boldsymbol{\mathcal{C}}$.\\
			Update mean $\boldsymbol{\mathcal{C}}_c$ of $\boldsymbol{\mathcal{C}}$ using (\ref{eq_cell_mean_incre}).\\
			Update covariance $\boldsymbol{\mathcal{C}}_{\Sigma}$ of  $\boldsymbol{\mathcal{C}}$ using (\ref{eq_cell_cov_incre}).
		}
	\end{algorithm}
	
	\section{2D histogram of rotation invariance}\label{sect_2d_histogram_ri}
	
	The main idea of our fast loop detection is that we use the 2D image-like histograms to roughly describe the keyframe. The 2D histogram describes the distribution of the Euler-angles of the feature direction in a keyframe.
	
	
	\subsection{The feature type and direction in a cell}
	
	As mentioned previously, each keyframe consists of a number of (e.g., 100) frames and each frame (i.e., scan) is partitioned into cells. For each cell, we determine the shape formed by its points and the associated feature direction (denoted as $\boldsymbol{\mathcal{C}}_{d}$). Similar to  \cite{magnusson2009appearance}, we perform eigenvalue decomposition on the covariance matrix $\boldsymbol{\mathcal{C}}_{\Sigma}$ in (\ref{eq_cell_cov}):
	
	\begin{equation}
	\label{eq_eig_decompose_sigma}
	\boldsymbol{\mathcal{C}}_{\Sigma}\mathbf{V} = \mathbf{V}\boldsymbol{\Lambda}
	\end{equation}
	where $\boldsymbol{\Lambda}$ is diagonal matrix with eigenvalues in descending order (i.e., $\lambda_1 \geq \lambda_2 \geq \lambda_3$). In practice, we only consider cells with 5 or more points to increase the robustness. 
	
	\begin{itemize}
		\item \textbf{\textit{Cell of plane shape}}: If $\lambda_2$ is significantly larger than $\lambda_3$, we regard this cell as a plane shape and regard the plane normal as the feature direction, i.e., $\boldsymbol{\mathcal{C}}_{d} = \mathbf{V}_3$ where $\mathbf{V}_3$ is the third column of the matrix $\mathbf{V}$. 
		\item \textbf{\textit{Cell of line shape}}: If the cell is not a plane and $\lambda_1$ is significantly larger than $\lambda_2$, we regard this cell as a line shape and regard the line direction as the feature direction, i.e., $\boldsymbol{\mathcal{C}}_{d} = \mathbf{V}_1$, the first column of $\mathbf{V}$. 
		\item \textbf{\textit{Cell with no feature}}: A cell which is neither a line nor plane shape is not considered.
	\end{itemize}
	
	\begin{figure}[t]
		\centering
		{\includegraphics[width=1\linewidth]{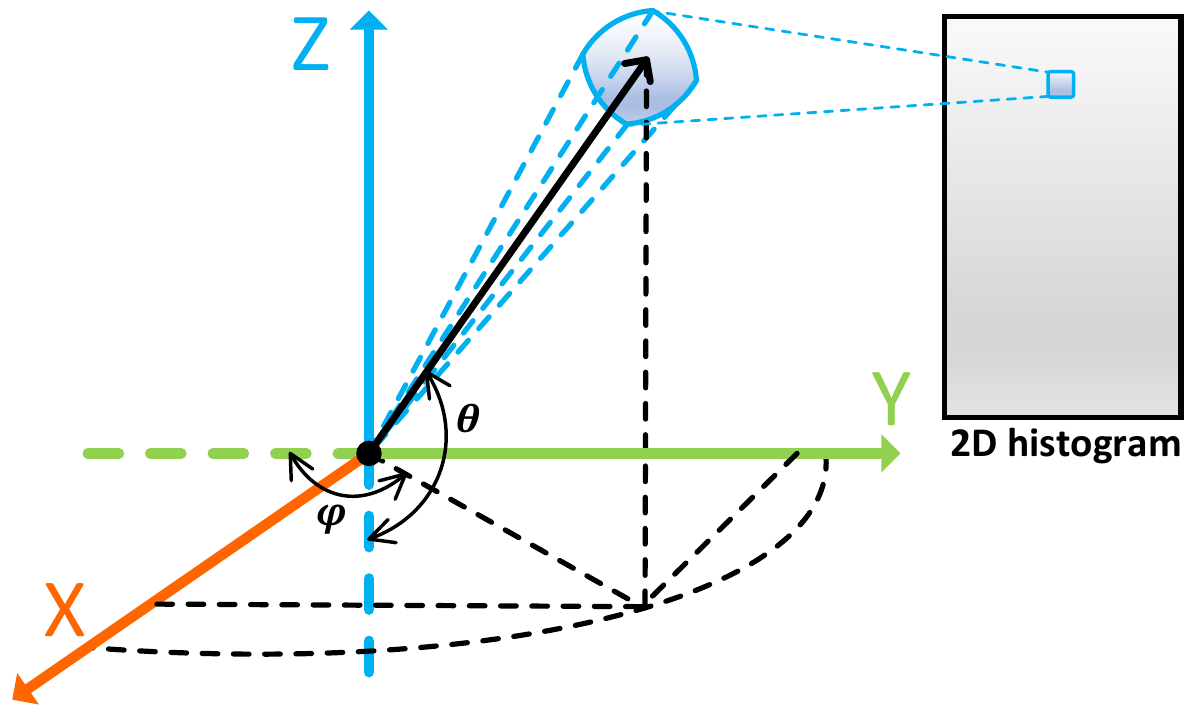}
			\caption{The Euler angle of a feature direction and its contribution to the 2D histogram, each element of the 2D histogram is the number of feature directions with pitch $\theta$ and yaw $\phi$ located in the corresponding bin.}	
			\label{fig_angle_histogram}}
		\vspace{-1.2cm}
	\end{figure}
	
	\subsection{Rotation invariance}\label{sect_rotaion_invariant}
	In order to make our feature descriptors invariant to arbitrary rotation of the keyframe, we rotate each feature direction $ \boldsymbol{\mathcal{C}}_{d} $ by multiplying it to an additional rotation matrix $\mathbf{R}$, and expect that most of the feature direction are lie on $X$-axis, and the secondary most are on $Y$-axis. Since plane feature is more reliable than line feature (e.g., the edge of plane feature are treated as a line feature), we use the feature direction of plane cells to determine the rotation matrix $\mathbf{R}$. Similar to the previous sections, we compute the covariance $\boldsymbol{\Sigma}_d$ of all plane feature directions in a keyframe:
	\begin{equation}
	\boldsymbol{\Sigma}_d = \sum_{i=1}^N  \boldsymbol{\mathcal{C}}_{i_d} \boldsymbol{\mathcal{C}}_{i_d}^T
	\end{equation}
	where $N$ is the number of plane cells, $\boldsymbol{\mathcal{C}}_{i_d}$ denotes the feature direction (i.e., plane normal) of the $i$-th plane cell. Similarly, the eigenvalue decomposition of $\boldsymbol{\Sigma}_d$ is:
	\begin{equation}
	\label{eq_eig_featrue_direct}
	\boldsymbol{\Sigma}_d\mathbf{V}_d = \mathbf{V}_d\boldsymbol{\Lambda}_d
	\end{equation}
	where $\boldsymbol{\Lambda}_d$ is a diagonal matrix with eigenvalues in descending order ($\lambda_{d_1} \geq \lambda_{d_2} \geq \lambda_{d_3}$), $\mathbf{V}_d = \begin{bmatrix} \mathbf{V}_{d_1} &  \mathbf{V}_{d_2} & \mathbf{V}_{d_3} \end{bmatrix}$ is the eigenvector matrix. Then, the rotation matrix $\mathbf{R}$ is determined as:
	\begin{equation}
	\mathbf{R} = \begin{bmatrix} \mathbf{V}_{d_1} &  \mathbf{V}_{d_2} &  \mathbf{V}_{d_1}\times \mathbf{V}_{d_2} \end{bmatrix}^T
	\end{equation}
	
	After compute the rotation matrix $\mathbf{R}$, we apply the rotation transformation to all feature (both plane and line) directions. 
	
	\begin{algorithm}[h]
		\caption{Computing the 2D hist. of a keyframe}
		\label{alg_2d_histogram}
		\renewcommand{\thealgocf}{}
		\renewcommand{\theAlgoLine}{}  
		\SetKwInOut{Input}{Input}
		\SetKwInOut{Output}{Output}
		\SetKwInOut{Begin}{Begin}
		\SetKwInOut{Start}{Start}
		\SetKwProg{Fn}{Function}{}\\
		\Input{Current keyframe  $\boldsymbol{\mathcal{F}}$ }
		\Output{2D Histogram $\mathbf{H_L}$ of line cell\\
			2D Histogram $\mathbf{H_P}$ of plane cell}
		\Start
		{	$\mathbf{H_L} \leftarrow \mathbf{0}$ , $\mathbf{H_P} \leftarrow \mathbf{0}$.\\
			Compute rotation matrix $\mathbf{R}$ from Sect.~\ref{sect_rotaion_invariant}.
		}
		
		
		\For{each $ \boldsymbol{\mathcal{C}} \in \boldsymbol{\mathcal{F}} $}
		{
			\If{ $ \boldsymbol{\mathcal{C}} $ is a line shape }
			{
				$\boldsymbol{\mathcal{C}}_d \leftarrow \mathbf{R}\boldsymbol{\mathcal{C}}_d$ \\
				Compute the pitch $\theta$ and yaw $\phi$ angle of $\boldsymbol{\mathcal{C}}_d$.\\
				$\mathbf{H_L}[ \text{round}(\phi/3^\circ), \text{round}(\theta/3^\circ) ]$ +=1 
			}
			
			\If{ $ \boldsymbol{\mathcal{C}} $ is a plane shape }
			{
				$\boldsymbol{\mathcal{C}}_d \leftarrow \mathbf{R}\boldsymbol{\mathcal{C}}_d$ \\
				Compute the pitch $\theta$ and yaw $\phi$ angle of $\boldsymbol{\mathcal{C}}_d$.\\
				$\mathbf{H_P}[ \text{floor}(\phi/3^\circ), \text{floor}(\theta/3^\circ) ]$ +=1 
			}
		}
		Gaussian blur $ \mathbf{H_P} $ and $ \mathbf{H_L} $. \\
	\end{algorithm}

	\begin{figure}[t]
		\centering
		\includegraphics[width=1\linewidth]{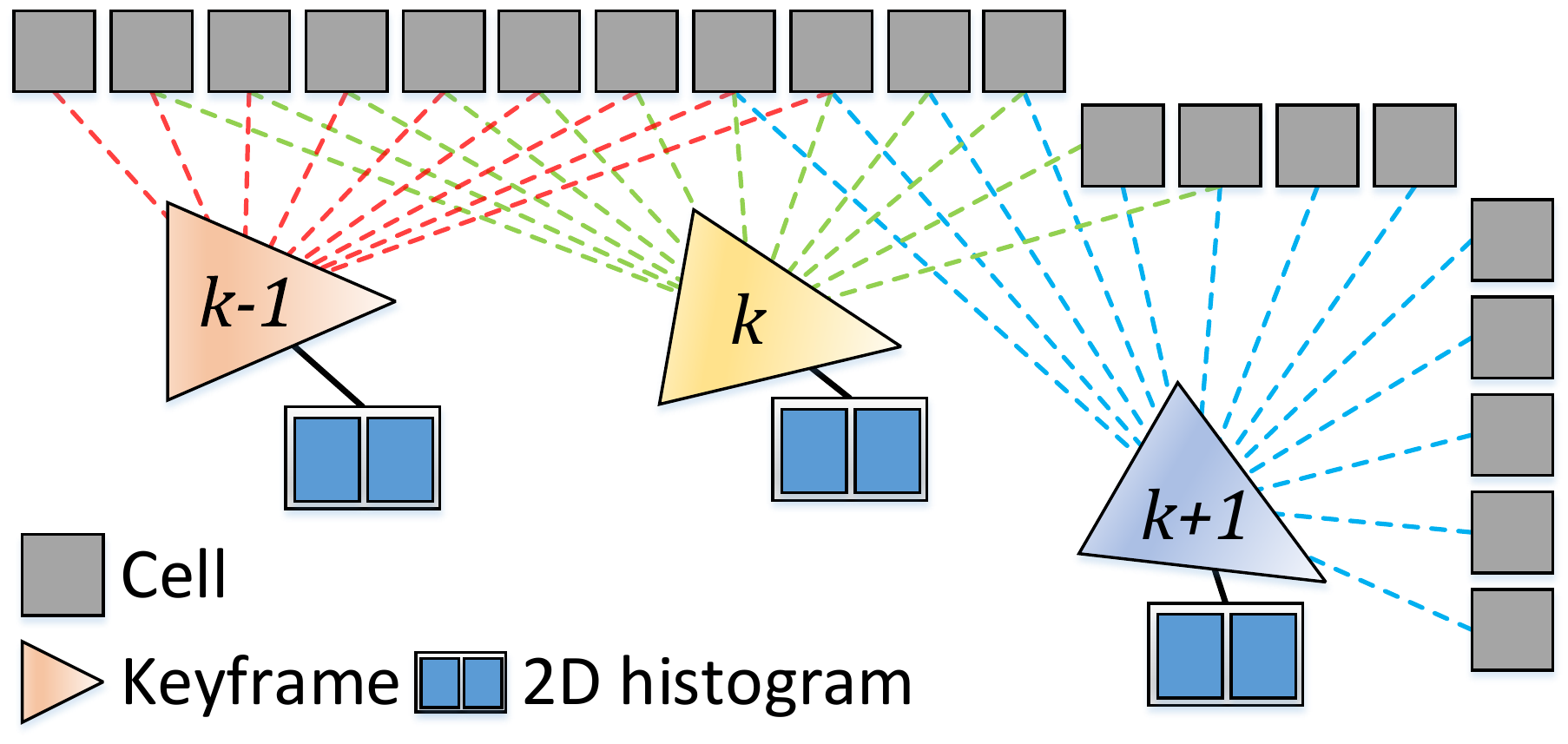}
		\caption{A keyframe is consists of $n$ (e.g., $n = 100$) frames (not shown), which then contains many cells. Each keyframe has two 2D histograms, one for line cells and the other one for plane cells.}
		\label{fig_fast_loop_detect}
		\vspace{-0.8cm}
	\end{figure}
	
	\subsection{2D histogram of keyframe}
	\begin{figure*}[htp]
		\subfigure[\label{fig_vis_map_cell_2d} The visualization of two keyframes (frame A and B), 2D histograms, and contained cells. Fig.~A1 is the RGB image of the first scene. Fig.~A2 and Fig.~A3 are the side-view and bird view of the keyframe, respectively. In  Fig.~A2 and Fig.~A3, the red points denote the raw point cloud in the keyframe, the white cubes denote the cells, the blue lines are the feature direction of plane cells, and the yellow lines are the feature direction of line cells. Fig.~A4 and Fig.~A5 are the 2D histogram (pixels are colored by their values) of plane and line features, respectively. The arrangement of  Fig.~B1$\sim$B5 is the same as Fig.~A1$\sim$A5.]
		{\includegraphics[width=1.40\columnwidth]{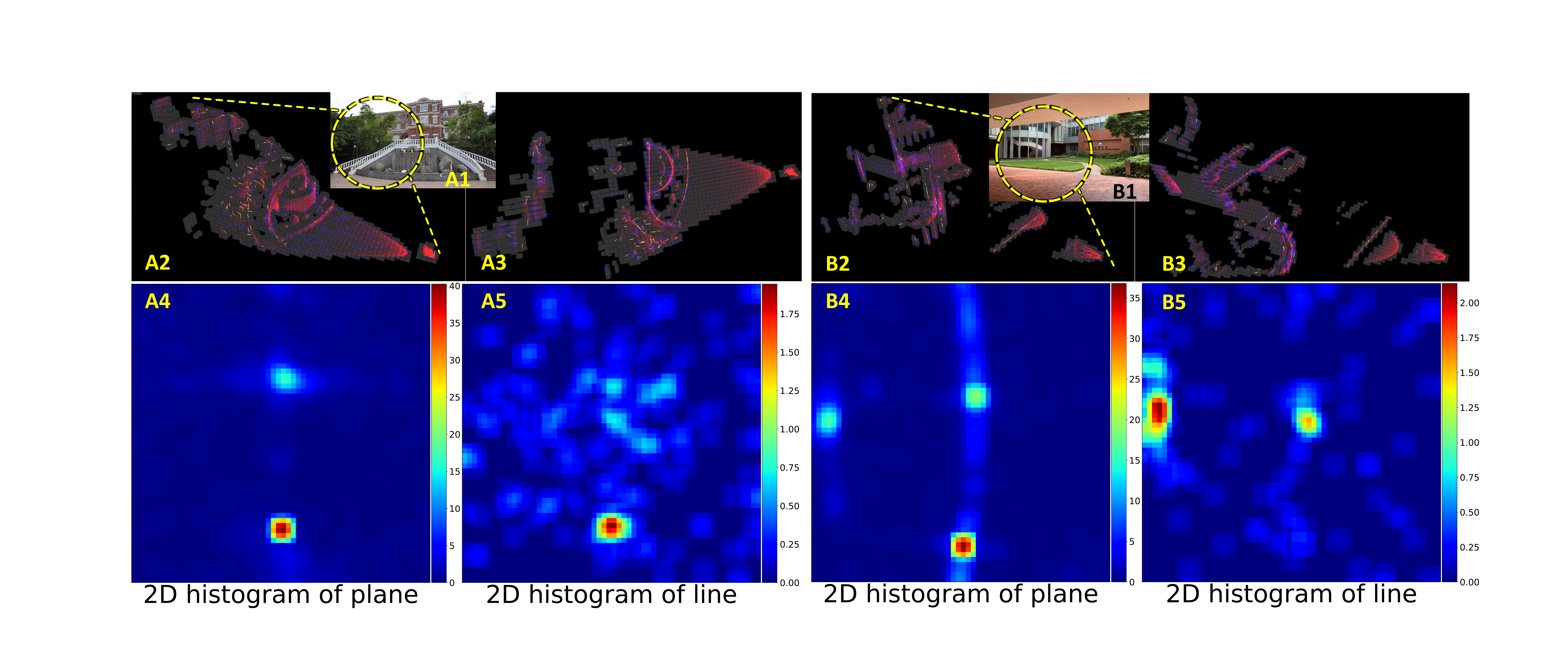}}
		\subfigure[\label{fig_rotationinvariant} The similarity in plane features between frame A and A ("P\_A2A"), and between frame A and B  ("P\_A2B"), and in line features between frame A and A ("L\_A2A"), and between frame A and B  ("L\_A2B"). Polar distance is the similarity level while ploar angle is the magnitude of random rotations. ]
		{\includegraphics[width=0.60\columnwidth]{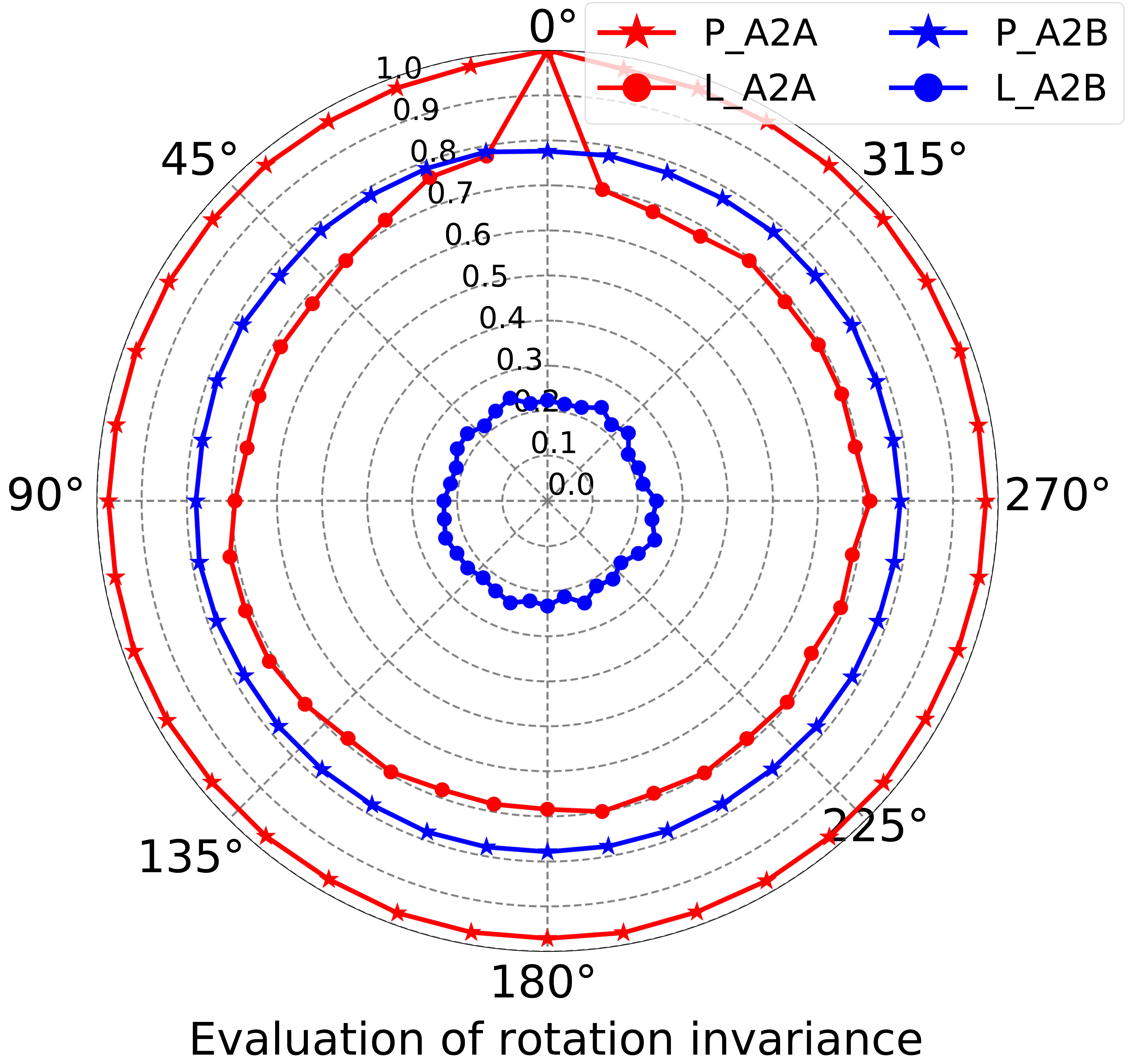}}
		\caption{The visualization of keyframe, cells, and 2D histograms (a) and the evaluation of rotation invariance (b).}
		\vspace{-0.5cm}
	\end{figure*}
	
	With the rotation invariant feature directions of all cells in a keyframe, we compute the 2D histogram as follows: 
	
	Firstly, for a given feature direction $\boldsymbol{\mathcal{C}}_{d} = [\boldsymbol{\mathcal{C}}_{d_x},\boldsymbol{\mathcal{C}}_{d_y},\boldsymbol{\mathcal{C}}_{d_z}]$, we choose the direction with positive $X$ components, i.e., $\boldsymbol{\mathcal{C}}_{d} =  sign(\boldsymbol{\mathcal{C}}_{d_x}) \cdot \boldsymbol{\mathcal{C}}_{d}$. Then, the Euler angle of the feature direction is computed (see Fig.~\ref{fig_angle_histogram}): 
	\begin{align}
	\theta &= \sin^{-1}{\left( \boldsymbol{\mathcal{C}}_{d_z}  \right) + 90^\circ } \in \left[0^{\circ}, 180^{\circ} \right] \\
	\phi &= \tan^{-1}{\left( \boldsymbol{\mathcal{C}}_{d_y} / \boldsymbol{\mathcal{C}}_{d_x} \right)  } + 90^\circ \in \left[0^{\circ}, 180^{\circ} \right]
	\end{align}
	
	The 2D-histogram we use is a $ 60\times 60 $ matrix (have $3^\circ$ resolution on both pitch and yaw angle), the elements of this matrix denote the number of line/plane cell with its pitch $\theta$ and yaw $\phi$ located in the corresponding bin. For example, $i$-th row, $j$-th column element, $e_{ij}$, is the number of cells with the angle of its feature direction satisfied:
	$$
	\begin{aligned}
	j\times 3^\circ\leq &\theta < (j+1)\times 3^\circ \\
	i\times 3^\circ\leq &\phi < (i+1)\times 3^\circ
	\end{aligned}
	$$ 

	To increase the robustness of the 2D histogram to possible noise, we apply a Gaussian blur on each 2D histogram we computed.
	
	The complete algorithm of computing the 2D histogram with rotation invariance is shown in Algorithm.~\ref{alg_2d_histogram}.

	\section{Fast loop detection}

	\subsection{Procedure of loop detection}
	
	As mentioned previously, we group $n$ frames (e.g., $n = 100$) into a keyframe $\boldsymbol{\mathcal{F}}$. It can be viewed as a local patch of the global map $\boldsymbol{\mathcal{M}}$, and contains all of the cells appearing in the last $n$ frames, as shown in Fig.~\ref{fig_fast_loop_detect}. We compute the 2D histogram of a new keyframe $\boldsymbol{\mathcal{F}}$ and its similarity (Section VI. B) with all keyframes in the past to detect a loop. The keyframe with a detected loop is then matched to the map (Section VI. C) and the map is updated with a pose graph optimization (Section VI. D). 
	
	\subsection{Similarity of two keyframes}
	For each newly added keyframe, we measure its similarity to all history keyframes.  In this work, we use the normalized cross-correlation of 2D histograms to compute their similarity, which has been widely used in the field of computer vision (e.g., template matching, image tracking, etc.). The similarity $S(\mathbf{H}_1, \mathbf{H}_2) $ of two 2D histogram $\mathbf{H}_1, \mathbf{H}_2$ is  computed as:
	\begin{equation}\label{eq_similarity_image}
	S(\mathbf{H}_1, \mathbf{H}_2) = \dfrac{ \sum_I( \mathbf{H}_1(I) - \bar{\mathbf{H}}_1 )( \mathbf{H}_2(I) - \bar{\mathbf{H}}_2 ) }{ \sqrt{\sum_I{\left( \mathbf{H}_1(I)- \bar{\mathbf{H}}_1\right)^2} \sum_I{\left( \mathbf{H}_2(I)- \bar{\mathbf{H}}_2\right)^2} }}
	\end{equation}
	where $\bar{\mathbf{H}}_k = \dfrac{1}{N}\sum_I \mathbf{H}_k(I) $  is the mean of $\mathbf{H}_k$ and $I = (i, j)$ is the index of the element in $\mathbf{H}_k$. If the similarity $S(\mathbf{H}_1, \mathbf{H}_2)$ between two keyframes is higher than threshold (e.g., $0.90$ for plane and $0.65$ for line), a loop is thought to be detected.

	\subsection{Maps alignment}
	
	After the successful detecting of a loop, we perform  maps alignment to compute the relative pose between two keyframes. The problem of maps alignment can be viewed as the registration between the target point cloud and source point cloud, as their work of \cite{pulli1999multiview}.
	
	Since we have classified the cell of linear shape and planar shape in our LOAM algorithm \cite{lin2019loam}, we use the feature of edge-to-edge and planar-to-planar to iteratively solve the relative pose.
	
	After the alignment, if the average distance of the points of edge/plane feature on is close enough to the edge/plane feature (distance less than $0.1 m$), we regard these two maps are aligned. 
	
	
	\subsection{Pose graph optimization}
	As the workflow is shown in Fig.~\ref{fig_workflow}, once the two keyframes are aligned, we perform the pose graph optimization following the method in \cite{olson2006fast, grisetti2010tutorial}. We implement the graph optimization  using the Google  \textit{ceres-solver}\footnote{\url{http://ceres-solver.org/}}. After optimizing the pose graph, we update all the cells in the entire map by recomputing the contained points, the points' mean and covariance.

	\begin{figure*}[htp]
		\includegraphics[width=2.00\columnwidth]{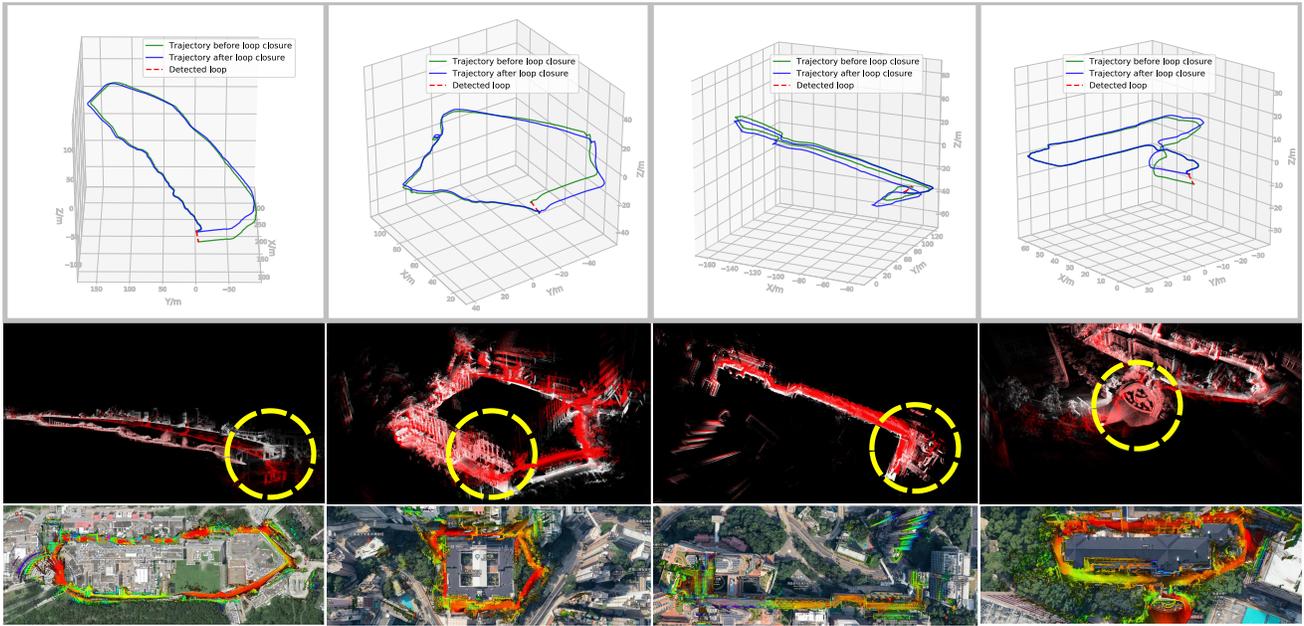}
		\caption{We test our algorithm on four datasets, which are all sampled by Livox-MID40. The first one is a large scale dataset  sampled in HKUST campus; The second one is sampled around a square building (main building of HKU); The third one is sampled indoor consisting of two long corridors in two neighboring floors. The fourth one is sampled around a rectangular building (Chong Yuet Ming Cultural Center in HKU) with many natural objects, such as trees, stairs, sculptures, etc.
		}
		\label{fig_dataset}
		\vspace{-0.5cm}
	\end{figure*}
	

	\section{Results}
	\subsection{Visualization of keyframe, cells, and 2D histograms }
	We visualize the two keyframes, their associated 2D histograms local maps, and contained cells in Fig.~\ref{fig_vis_map_cell_2d}. This figure shows that the 2D histogram of the two different scenes are very distinctive.
	
	\subsection{Rotation invariance}
	We evaluate the rotation invariance of our loop detection method by computing the similarity of the two scenes in Fig.~\ref{fig_vis_map_cell_2d} with random rotations. For each level of rotation, we generate 50 random rotation matrix of random directions but the same magnitude, rotate one of the two scenes by the generated rotation matrix, and compute the average similarity among all the 50 rotations of the same magnitude. The similarity of keyframe A to itself, keyframe B to itself, and keyframe A to keyframe B are shown as Fig.~\ref{fig_rotationinvariant}. It can be seen that, the similarity of plane features almost hold the same under different or rotation  magnitude, and the similarity of the  same keyframe (with arbitrary rotation) is constantly higher than the similarity of different keyframes. For line features, although the similarity of the same keyframe slightly drops when rotation takes place, it is still significantly higher than the similarity of different keyframes.
	
	\subsection{Time of computation}
	We evaluate the time consumption or each step of our system on two platforms: the desktop PC (with \textit{i7-9700K}) and onboard-computer (DJI manifold2\footnote{\url{https://www.dji.com/cn/manifold-2}} with \textit{i7-8550U}). The averge running time of our algorithm run on \textit{HKUST large scale dataset} (the first column of Fig.~\ref{fig_dataset}) are shown in Table.~\ref{tab_time_profile}, where we can see our proposed method is fast and suitable for real time scenenarios on both platforms.
	
	\begin{table}
		{
			\setlength{\extrarowheight}{.1em}
			\setlength\extrarowheight{0.01pt}
			\begin{tabular}[h]{|c|c|c|c|}
				\hline
				& 2D histogram & Maps  & Similarity of   \\
				& computing  & alignment & two maps  \\
				\hline
				Desktop PC& 1.18 $ ms $ & 621 $ ms $ & 13 $\mu s$ \\
				\hline
				Onboard-computer & 1.48 $ ms $ & 931 $ ms $ & 16 $\mu s$ \\
				\hline
			\end{tabular}
			\caption{The time table or our system run on two platforms.}
			\label{tab_time_profile}
			\vspace{-0.8cm}
		}
	\end{table}
	
	
	\subsection{Large scale loop closure results}
	We test our algorithm on four datasets in Fig. \ref{fig_dataset}, where the first row is the comparison of trajectory before (green solid line) and after (blue sold line) loop closure, the red dashed lines indicate the detected loop. The second row of figures is the comparison of the point cloud map before (red) and after loop closure (white), where we can see the loop closure can effectively reduce the drift of LiDAR odometry and mapping (especially in the areas inside yellow circle). We align our point cloud after loop closure with Goolge maps in the third row, where we can see the alignment is very accurate, showing that the accuracy of our system is of high precision. 
	
	\section{Conclusion}
	
	
	This paper presented a fast, complete, point cloud based loop closure for LiDAR odometry and mapping, we develop a loop detection method which can quickly evaluate the similarity of two keyframes from the 2D histogram of plane and line features in the keyframe. We test our system on four datasets and the results demonstrate that our loop closure can significantly reduce the long-term drift error. We open sourced all of our datasets and codes on Github to serve as an available solution and paradigm for point cloud based loop closure research in the future. 
	\bibliography{icra2020loop}
	
\end{document}